# A Comprehensive Evaluation of Graph Neural Networks and Physics-Informed Learning for Surrogate Modelling of Finite Element Analysis

Nayan Kumar Singh

Independent Researcher, Bangalore, India

Email: nayan.ksingh.r@gmail.com

___

**Abstract**

Although Finite Element Analysis (FEA) is an integral part of the product design lifecycle, the analysis is computationally expensive, making it unsuitable for many design optimization problems. The deep learning models can be a great solution. However, selecting the architecture that emulates the FEA with great accuracy is a challenge. This paper presents a comprehensive evaluation of graph neural networks (GNNs) and 3D U-Nets as surrogates for FEA of parametric I-beams. We introduce a Physics-Informed Neural Network (PINN) framework, governed by the Navier-Cauchy equations, to enforce physical laws. Crucially, we demonstrate that a curriculum learning strategy—pre-training on data followed by physics-informed fine-tuning—is essential for stabilizing training. Our results show that GNNs fundamentally outperform the U-Net. Even the worst performer among GNNs, the GCN framework, achieved a relative L2 error of 8.7% while the best framework among U Net, U Net with attention mechanism trained on high resolution data, achieved 13.0% score. Among the graph-based architectures, the Message Passing Neural Networks (MPNN) and Graph Transformers achieved the highest accuracy, achieving a relative L2 score of 3.5% and 2.6% respectively. The inclusion of physics fundamental laws (PINN) significantly improved the generalization, reducing error by up to 11.3% on high-signal tasks. While the Graph Transformer is the most accurate model, it is more 37.5% slower during inference when compared to second best model, MPNN-PINN. The PINN-enhanced MPNN (MPNN-PINN) provides the most practical solution. It offers a good compromise between predictive performance, model size, and inference speed.

## 1. Introduction

Finite Element Analysis (FEA) has slowly replaced the traditional design-test cycles to design-simulation-test cycle, thereby reducing the prototyping and testing cost and significantly improving the product design timelines. FEA has shown tremendous applications in structural and thermal analysis. While FEA is still a powerful tool, the time required for meshing, run and post-processing makes it difficult to use it for real time applications like digital twins or design optimization problems requiring multiple runs. The high computation cost has forced engineers to look for efficient but simpler alternative models.

For these types of problem, historically, engineers have used efficient statistical representative techniques such as Reduced Order Models (ROM), for e.g. Proper Orthogonal Decomposition (POD), Response Surface Methods and Kriging.[1, 2] These techniques tries to reduce the solution space dimensionality by projecting the it to a lower dimension while trying to retain as much as relevant data as possible and discarding the noise. This allows for quick turnaround time for generating new solutions.[3] However, this unmatched efficiency comes with limitations in accuracy. One of the primary assumptions while building these models is linearity. Thus, their effectiveness on highly non - linear phenomena is questionable. Furthermore, the integration of methods like POD-Galerkin is often impractical due to "intrusive" requirements. They require modification of the solver code, which is usually not feasible in commercial FEA software. These models also struggle to generalize to designs that lie far from the initial training data. The models also behave poorly when data is by high-dimensional, dimensionality can't be reduced without compromising on relevancy.[4]

These challenges have motivated the search for more flexible, non-linear, and non-intrusive methods. Deep learning-based surrogate models have emerged as a promising solution satisfying the requirements. These models are when trained on input-output mapping generated from FEA solvers provide inference on unseen data in time that is order of magnitudes lesser than the FEA solver. The FEA models work via numerical methods applied to structure of nodes and elements, unstructured mesh. Thus, Graph Neural Networks (GNNs) offer a particularly powerful inductive bias.[5] By representing the FEA mesh as a graph, GNNs can leverage the existing architecture, allowing for a more natural and efficient learning. Thus GNNs should be the natural choice to learn FEA simulated physical phenomena like stress and strain propagation

compared to grid-based methods like Convolutional Neural Networks (CNNs).[6]

While the DL surrogate models looks promising, they come with their own challenges that needs to be addressed. Similar to the ROM models purely data-driven models may struggle to generalize to out-of-distribution scenarios. In these cases, they can produce physically implausible results, as they lack any knowledge of the fundamental system's governing laws. The limitation can be easily fixed by simply telling the model the fundamental laws. This can be done by embedding the governing partial differential equations (PDEs) directly into the neural network's loss function.[7][8] This approach is called as Physics informed neural network (PINN). This method works as a strong regularizer that guides the model toward a solution that is not only accurate with respect to the training data but is also consistent with fundamental physical principles.[9]

The goal of this study is to develop GNN based surrogate models imitating the FEA solver for deformation analysis of I Beam and improve the model via PINN integration. The key contributions are as following:

1. **Architectural Comparison:** Multiple GNN architectures - GCN, GAT, MPNN, Graph Transformer - are compared to a 3D U-Net baseline on different dataset – low input signal vs high input signal, multimodal vs unimodal load distributions.

2. **Successful Physics Embedment in GNNs for deformation analysis - PINN:** The Navier-Cauchy equation is successfully integrated into the GNN training process. Thereby linear elasticity fundamental laws are indirectly told to the model to significantly improve the model generalization.

3. **A robust PINN training strategy:** Curriculum learning technique has been introduced and validated. This method used physics loss weight annealing. The method proves to be critical for training and subsequent convergence of PINN models.

4. **Performance-efficiency analysis:** The graph transformer has been identified as the most accurate architecture, however the PINN-enhanced MPNN (MPNN-PINN) proves to be a superior solution for practical deployment in real-time applications - an optimal balance of predictive accuracy, model size, and inference time.

## 2. Related Work

The research is focused on three topics – deep learning for FEA simulation, mesh-based Graph Neural Networks, and Physics-Informed Machine Learning.

### 2.1. Deep Learning Surrogates for Physical Simulation

The use of deep learning to emulate the simulations is an emerging field of research that shows good potential for practical implementations. Early on, the research was focused on using standard architectures like Multi-Layer Perceptrons (MLPs) for low-dimensional feature space or Convolutional Neural Networks (CNNs) for problems defined on regular, grid-like domains, usually seen in computational fluid dynamics. However, structural analysis and solid mechanics simulations are typically done on unstructured mesh.[6] This created a need to develop methods and architectures that can handle such irregularities in numerical domain.

### 2.2. Mesh-Based Graph Neural Networks in Mechanics

Graph Neural Networks have recently gained popularity for learning on mesh-based data. [5, 6] The FEA simulation is performed on a network on nodes with elements connecting them. This structure is directly given to the GNN model. This "additional knowledge" gives an edge to GNNs over other modelling techniques. The GNNs at their core is "silently" emulating the FEA models – message passing is similar to the numerical physical data flow between nodes. Pfaff et al. work on MeshGraphNets demonstrates the ability of GNNs to simulate a wide variety of physical systems defined on unstructured meshes.[10] While these works establishes GNNs as a natural choice to work on unstructured mesh problem, a systematic comparison of different GNN architectures on complex, multi-modal structural mechanics tasks is less explored. Our work contributes a rigorous, comparative study to identify the most effective architectures for this domain.

### 2.3. Physics-Informed Neural Networks (PINNs)

The concept of embedding physical laws into neural networks was formalized by Raissi, Perdikaris, and Karniadakis, who introduced Physics-Informed Neural Networks (PINNs).[11][12] PINNs augment the standard data-driven loss function with a second term that penalizes deviations from the governing Partial Differential Equations (PDEs).[7] This physics-based loss is calculated on a set of collocation points and is typically evaluated using automatic differentiation to compute the necessary derivatives. By training to minimize this composite loss, the network is constrained to learn solutions that are physically consistent.[8] Our work applies this paradigm not to solve the PDE from scratch, but as a physics-based regularizer to improve the generalization of an already powerful data-driven GNN surrogate.

### 2.4. Challenges and Advances in PINN Training

Despite their potential, training PINNs can be notoriously challenging, often suffering from instability or slow convergence. A key difficulty lies in balancing the gradients from the data-driven loss and the various terms of the physics-based loss. To address this, several advanced training strategies have been proposed, including adaptive weighting schemes and curriculum learning.[13] Curriculum learning, where the model is exposed to progressively harder tasks, is particularly promising as a method to improve convergence and stability.[14][15] Our work contributes to this area by demonstrating a specific, robust curriculum strategy—pre-training and fine-tuning with loss weight annealing—and proving its necessity and effectiveness for stabilizing the training of a GNN-based PINN for a complex structural mechanics problem.

## 3. Methodology

### 3.1. Problem Formulation and Data Generation

I beam was chosen as the element for study as this geometry is simple enough to parameterize and generate different datasets but it's anisotropic bending stiffness is a non-trivial problem for learning. For all the ground truth data, the geometry and the mesh were created using open source gmsh; and the problem was processed in the DOLFINx FEA solver.

For better generalization, three domains were considered for parameterization – geometry, material properties and the loading condition, for more details refer Table 1. Latin Hypercube Sampling (LHS) was used for sampling for efficient and uniform space exploration.

Mesh element size was intentionally kept constant for the whole ground truth dataset. Along with it the number of nodes and their connectivity remained constant for the complete dataset. However, the node coordinates were updated to accommodate geometric variation. This was done to ensure a consistent graph structure. This enabled the model to learn the underlying physics, the impact of changing geometric, material, and load parameters, without confusing the model with changing mesh discretization. Solver with finer mesh typically generates much accurate result that is consistent with physics. All the pre-processing was done to generate best possible ground truth that is practical with time and computation resources available. Thus, we get a consistent basis for comparison across all models, particularly the GNNs that operate on this graph structure.

**Dataset Generation: Low Signal vs. High Signal Regimes**

It is well known that FEA solvers perform relatively poorly on signal datasets due to higher signal to noise ratio (SNR). One of the primary goals for this study is to find out if the DL based surrogate models are capable to distinguish numerical noise in ground truth from physical outcome. The higher signal (load) also produces larger label (displacement). Thus, learning input to output mapping is easier. In this study we explore the capability of DL models trained on low signal dataset to generalise on high signal ground truth. Thus, two different datasets were generated:

1. **Low Signal Dataset:** This dataset consists of 1500 simulations with force within range of 50kN to 100kN. The data contains a random mixture of all three load types (bending – along both weak and strong axis, and torsion).

2. **High Signal Dataset:** This dataset consists of 1000 simulations with force within range of 200kN to 250kN. This dataset only contains load responsible for bending along the string axis.

Table 1: Parametric space for the I-beam FEA simulations.

| Parameter | Description | Type | Range / Values |
|---|---|---|---|
| Beam Length | The length of the beam along the Z-axis. | Continuous | 280.0 – 320.0 mm |
| Flange Width | The total width of the top and bottom flanges. | Continuous | 90.0 – 110.0 mm |
| Flange Thickness | The thickness of the flanges. | Continuous | 13.0 – 17.0 mm |
| Web Thickness | The thickness of the central vertical web. | Continuous | 8.0 – 12.0 mm |
| Beam Depth | The total height of the I-beam. | Continuous | 140.0 – 160.0 mm |
| Fillet Radius | The radius of the fillets at the web-flange junctions. | Continuous | 10.0 – 14.0 mm |
| Youngs Modulus | The Young's Modulus of the material (variations of steel). | Continuous | GPa |
| Poisons ratio | The Poisson's Ratio of the material. | Continuous | 0.28 – 0.32 |
| Force Magnitude | The total magnitude of the force applied to the free end. | Continuous | kN |
| Load Type | The nature of the applied load. | Categorical | Bending y, bending x, torsion |
| Load Distribution | The spatial distribution of the applied load. | Categorical | Uniform, Linear Y |

**Task Formulation:** The two datasets enable the study of three different kind of models.

- **Generalist (Low Signal) Task:** This is the ultimate test of DL modelling capabilities. The model needs to differentiate noise from physics induced response. Also, the model needs to learn stress-strain response at more fundamental level as the type of loads are also varying. However, this can be advantageous for generalisation to an unseen type of geometry.

- **Specialist (Low Signal) Task:** This model is trained on data with bending along strong axis. However, data consists of low signal inputs.

- **Specialist (High Signal) Task:** This data should be easiest for DL surrogate model to learn as the model is trained on consistent type of loading with significant label magnitudes. Thus, ideally this should represent the best capability of DL surrogate models, providing a benchmark for the best possible performance on this specific case.

**Boundary and Loading Conditions:** The simulations model a cantilever beam configuration, as depicted in Figure 1. One end of the beam (at Z=0) is fully fixed, representing a clamped boundary condition. A distributed traction force is applied to the surface at the free end (at Z=L). The nature of this force is determined by the load type parameter: bending Y (vertical), bending X (horizontal), or torsion (twisting moment).

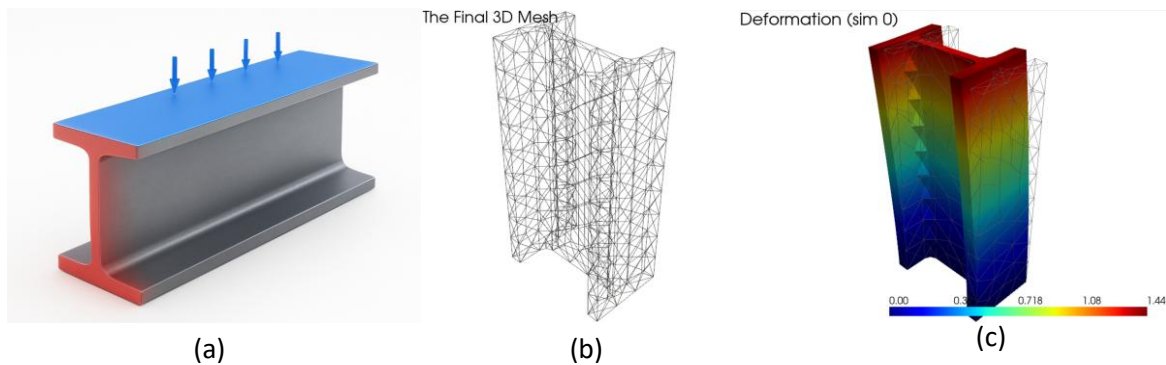

(a)　　　　　　　　　(b)　　　　　　　　　(c)

*Figure 1: FEA Problem Formulation and Sample Data.*
(a) The I-beam geometry with boundary conditions, showing the fixed surface (red) and the load application surface (blue). (b) A visualization of the unstructured tetrahedral mesh used for the FEA simulations. (c) A sample ground truth displacement field for a bending Y load case from the High Signal dataset, showing the magnitude of displacement.

## 3.2. Data Representation and Preprocessing

The high dimensional outputs from the FEA solver such as stress and displacement are represented by the pointwise values that are stored the nodes of the mesh. All data like mesh structure, displacement field and the input such as material properties was saved to a dedicated HDF5 (.h5) file. This approach preserves the FEA ground truth as much as possible. Thus, this collection of files serves as ground truth for GNN models. The other methods for data storage and pre-processing like element-wise averages would involve approximation resulting in loss of resolution. 3D Cartesian coordinate system (X, Y, Z) is used as the choice measurement of position.

While unstructured HDF5 data is suitable for GNNs, grid-based methods like 3D U-Net architecture needs a structured, voxelized input. To meet this requirement, a preprocessing pipeline was implemented to convert the unstructured data into a uniform grid representation. This process involves:

1. Grid Definition: A consistent boundary was used for all simulations to enable spatial alignment. Two type of grid resolution was used for this study - a low-resolution grid of 64 x 32 x 32 voxels and a high-resolution grid of 96 x 48 x 48 voxels. The two datasets enables us to study the impact of input resolution on model performance.

2. Field Interpolation: Trilinear interpolation was used to represent the unstructured displacement vectors onto regular grids. Voxels outside the geometry of original mesh were assigned zero value.

3. Geometry Mask Creation: A binary geometry mask was created using nearest-neighbour interpolation to inform the U-Net of the beam's location within the voxel space.

All input and output data were normalized to ensure stable and efficient training. Deep learning models are sensitive to the scale of input features. Large-valued parameters such as Young's Modulus could dominate the learning process, leading to unstable gradients. Similarly, normalizing the label (displacement) sets an appropriate scale for the loss function. Thus, all scalar input parameters and the output displacement fields were scaled to a range of approximately [-1, 1] using min-max scaling. This scaling was performed using the global minimum and maximum values observed across the entire training dataset. The same scaling factors were used for inverse transformation during inference. This is required to prevent data leak in test set and to return predictions in original physical units.

## 3.3. Model Architectures

To systematically evaluate the most effective approach for learning FEA surrogates, we implemented and compared two distinct classes of neural network architectures: grid-based Convolutional Neural Networks (CNNs) and mesh-based Graph Neural Networks (GNNs). Each class contains several variants to allow for a thorough analysis of performance, efficiency, and the impact of specific architectural features.

### 3.3.1. Grid-Based Architecture: 3D U-Net

To provide a strong baseline from the convolutional domain, we adapted the U-Net architecture to our 3D regression problem. The U-Net's encoder-decoder structure with skip connections, illustrated in **Figure 2**, is well-suited for capturing both local features and global context, which is essential for predicting a full displacement field.

- **Core Architecture:** As shown in Figure 2, our 3D U-Net consists of a contracting path (encoder) and an expansive path (decoder). Skip connections concatenate feature maps from the encoder to the corresponding layers in the decoder, which is crucial for preserving high-frequency details.

- **Input Formulation:** The input to the U-Net is a multi-channel 3D tensor. The first channel is the binary geometry mask, which explicitly defines the shape of the I-beam within the voxel grid. Subsequent channels are created by broadcasting each of the normalized scalar simulation parameters (e.g., force magnitude, Young's modulus, flange width) into its own full-resolution 3D channel. This "parameter embedding" technique ensures that every convolutional filter at every location has access to the global physical context of the simulation.

- **Architectural Variants:** As shown in our results, we evaluated two main variants based on the code in unet_variants.py:

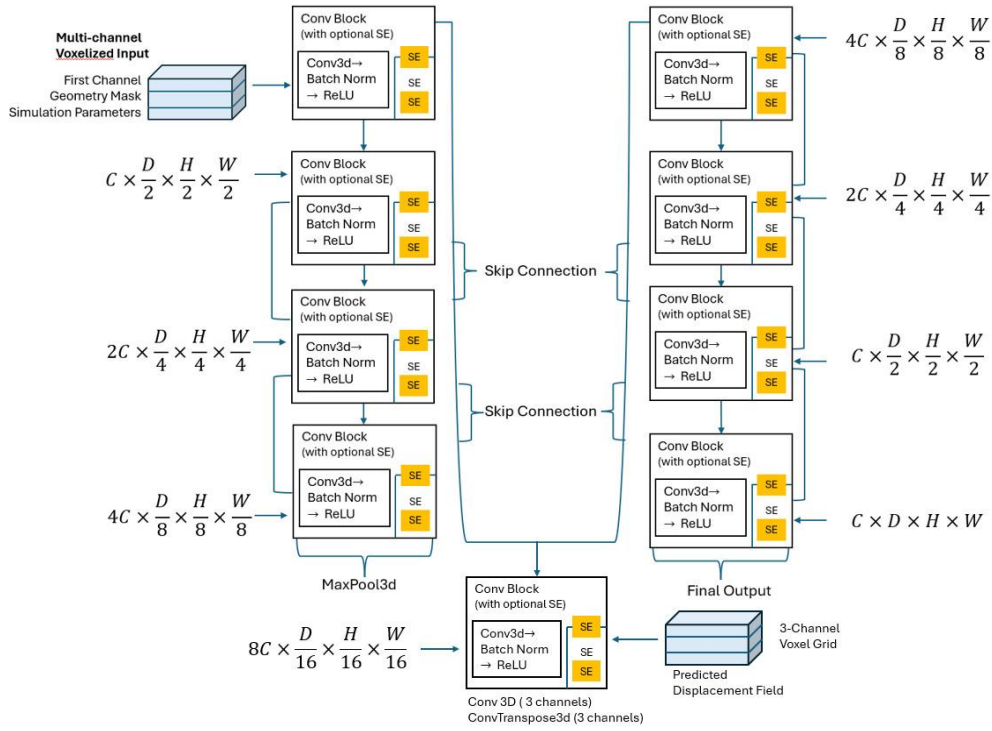

*Figure 2: The 3D U-Net Architecture. The model takes a multi-channel voxelized input, where the first channel is the geometry mask and subsequent channels are broadcasted simulation parameters. The encoder (left) progressively downsamples the spatial resolution while increasing feature depth. The decoder (right) symmetrically upsamples the features, using skip connections (grey arrows) to re-introduce high-resolution information from the encoder path. Optional Squeeze-and-Excitation (SE) blocks provide channel-wise attention within each convolutional block. The final output is a 3-channel voxel grid representing the predicted displacement field.*

1. **UNet3D:** A computationally efficient model with a baseline channel count of 32 in the first layer, which doubles with each downsampling step.

2. **Attention-Enhanced U-Net (UNet3D + Attn):** To test the hypothesis that focusing on salient features can improve performance, we integrated a **Squeeze-and-Excitation (SE) block** into each convolutional layer. The SE_Block3D is a channel-wise attention mechanism that adaptively recalibrates the feature maps. It "squeezes" global spatial information into a channel descriptor and then uses this to compute channel-wise attention weights, effectively allowing the network to emphasize more informative feature channels and suppress less useful ones.

### 3.3.2. Mesh-Based Architectures: Graph Neural Networks

GNNs represent a more natural paradigm for this problem, as they operate directly on the unstructured FEA mesh, thereby preserving the exact geometry and topology without any discretization error from voxelization. The general GNN paradigm we employ is shown in **Figure 3**.

- **Graph Representation:** The FEA mesh, composed of tetrahedral elements, was converted into an undirected graph structure suitable for PyTorch Geometric. The nodes of the graph directly correspond to the nodes of the FEA mesh. The graph's edges are derived by extracting all unique edges from the tetrahedral elements.

- **Node Feature Engineering:** Each node in the graph is initialized with a feature vector that encodes both its local position and the global context of the simulation. This vector is constructed by concatenating:

  1. The node's 3D Cartesian coordinates (pos).

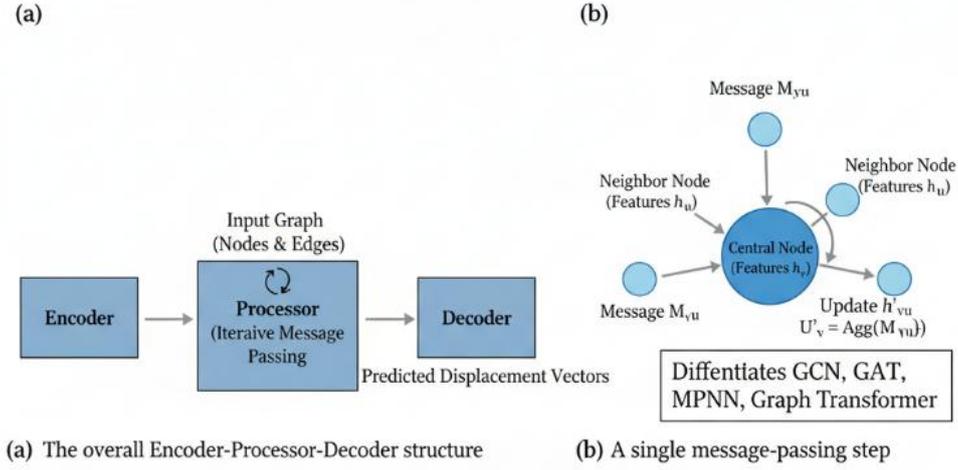

*Figure 3: The Graph Neural Network Paradigm.* (a) The overall Encoder-Processor-Decoder structure. The input graph's node features are encoded into a latent space, iteratively refined through multiple message-passing layers in the processor, and finally decoded into the predicted displacement vectors. (b) A conceptual view of a single message-passing step, where a central node aggregates information from its neighbours to update its own feature representation. The specific mathematical formulation of this aggregation and update step is what differentiates the GCN, GAT, MPNN, and Graph Transformer architectures.

2. The full set of normalized scalar simulation parameters, which are repeated for every node.

3. A **conditional load-type encoding**: For the *Generalist (multimodal)* model, the categorical load type is one-hot encoded into a 3-dimensional vector (e.g., [ 1, 0, 0] for bending Y). For the *Specialist (unimodal)* models, it is encoded as a single scalar. This distinction is critical, as the one-hot encoding provides a clear, non-ordinal signal that allows the generalist model to effectively learn the different physical responses.

- **Architectural Variants:** All GNNs follow the encoder-processor-decoder design shown in Figure 3a. An input linear layer encodes the node features into a higher-dimensional hidden state. A series of "processor" layers then perform message passing, illustrated in Figure 3b, to iteratively update these hidden states. Finally, a linear decoder maps the final hidden states to the predicted 3D displacement vectors. We evaluated four GNN processor types:

    1. **GCN (Graph Convolutional Network):** Uses GCNConv layers, which perform isotropic aggregation by averaging the features of neighbouring nodes. It serves as a foundational GNN baseline. Mathematically,

    $$h_v^{(l+1)} = \sigma\left( W^{(l)} \sum_{u \in \mathcal{N}(v) \cup \{v\}} \frac{1}{c_{vu}} h_u^{(l)} \right)$$

    Where;
    $h_v^{(l)}$: hidden representation of node $v$ at layer $l$
    $\mathcal{N}(v)$: neighbors of node $v$
    $c_{vu}$: normalization constant for the edge between $v$ and $u$
    $W^{(l)}$: weight matrix at layer $l$
    $\sigma(\cdot)$: activation function (e.g., ReLU)

    2. **GAT (Graph Attention Network):** Employs GATConv layers, which enhance GCN by introducing a self-attention mechanism. This allows the model to learn different weights for different neighbours, focusing on the most relevant information during aggregation. Mathematically,

    $$h_v^{(l+1)} = \sigma\left( \sum_{u \in \mathcal{N}(v) \cup \{v\}} \alpha_{vu} W^{(l)} h_u^{(l)} \right)$$

    The attention coefficients $\alpha_{vu}$ are computed using a learnable attention mechanism, allowing for a weighted, anisotropic aggregation.

3. **MPNN (Message Passing Neural Network):**

    Implemented using the expressive Meta Layer framework. This provides a more general form of message passing where separate neural networks (Edge Model and Node Model) are explicitly learned to first create "messages" based on pairs of connected nodes, and then update each node based on the sum of its incoming messages. Residual connections are used after each update to improve gradient flow. This provides a more general and expressive form of message passing by using distinct learnable functions (MLPs) for message creation (ψ) and node updates (φ). Mathematically,

    $$m_{vu} = \psi\left(h_v^{(l)}, h_u^{(l)}\right)$$

    $$h_v^{(l+1)} = \phi\left(h_v^{(l)}, \sum_{u \in \mathcal{N}(v)} m_{vu}\right)$$

4. **Graph Transformer:**
    Utilizes Transformer Conv layers, representing the most powerful architecture in our study. This layer applies multi-head self-attention to the local neighbourhood of each node, allowing it to learn highly complex and adaptive aggregation functions, capturing intricate dependencies between nodes.

### 3.4. Physics-Informed Learning Framework

To move beyond a purely data-driven approach and embed physical knowledge into our models, we integrated a Physics-Informed Neural Network (PINN) framework. The primary goal of the PINN component is not to solve the PDE from scratch, but rather to act as a physics-based regularizer, ensuring the model's predictions adhere to the governing laws of solid mechanics and thereby improving generalization.

#### 3.4.1. Governing Equations and Loss Formulation
The physical behaviour of a linearly elastic, isotropic solid in static equilibrium is governed by the **Navier-Cauchy equations**. In vector form, the equation is:

$$\mu \nabla^2 u + (\mu + \lambda)\nabla(\nabla \cdot u) + F = 0$$

where **u** is the displacement vector field, **F** is the body force vector (assumed to be zero in our case), and $\mu$ and $\lambda$ are the material-specific Lamé parameters, which are derived from the Young's Modulus and Poisson's Ratio.

Our total loss function is a composite of a data-driven term and a physics-based term, weighted by a dynamic parameter α(t):

$$L_{total} = L_{data} + \alpha(t) \times L_{physics}$$

- **Data Loss ($L_{data}$):** This is the Mean Squared Error (MSE) between the GNN's predicted displacement vectors at the mesh nodes and the ground truth displacements from the FEA solver. This term ensures the model remains faithful to the simulation data.

- **Physics Loss ($L_{physics}$):** This term quantifies the extent to which the model's predictions violate the Navier-Cauchy equations. It is calculated as the mean squared residual of the governing PDE over a large set of collocation points sampled randomly throughout the beam's volume at each training step. Crucially, all spatial derivatives required to compute the PDE residual (e.g., ∇u, ∇²u) are calculated analytically using **automatic differentiation**. This is a key advantage of using neural networks, as it allows us to approximate the differential operators with high precision by differentiating the network's output with respect to its input spatial coordinates.

#### 3.4.2. Treatment of Boundary Conditions
In many "classic" PINN applications that solve PDEs from scratch, an explicit boundary condition loss term ($L_{bc}$) is required. However, in our surrogate modelling framework, this is unnecessary. The Dirichlet boundary conditions (i.e., the zero-displacement constraint at the fixed end of the beam) are already present in the ground truth data. By training the model to minimize $L_{data}$, it implicitly learns to satisfy these boundary conditions. The physics loss $L_{physics}$ then regularizes the solution *within* the domain, conditioned on these data-enforced boundaries.

### 3.5. Curriculum Learning for PINN Stabilization

A significant challenge in training PINNs is balancing the gradients from the data and physics loss terms. Our initial attempts to train the PINN-enhanced GNNs with a fixed, non-zero weight α from the beginning of training proved to be unstable.

### 3.5.1. Observed Instability

The naive joint-training approach consistently failed to converge to a meaningful solution. The typical failure mode observed was a decreasing training loss while the validation loss either fluctuated erratically or steadily increased. This behaviour indicates that the optimizer was struggling with conflicting or poorly scaled gradients from the $L_{data}$ and $L_{physics}$ terms. The high-frequency nature of the second-order derivatives in the physics loss can easily dominate the training process in early stages before the model has learned a reasonable approximation of the solution, preventing the model from learning the fundamental input-output mapping.

### 3.5.2. The Successful Curriculum Strategy: Fine-Tuning with Annealing

To overcome this instability, we developed a robust two-stage curriculum learning strategy, reframing the task from joint training to **pre-training and fine-tuning**:

1. **Stage 1: Data-Driven Pre-training.** First, a GNN model is trained to convergence on the dataset using *only* the data loss ($L_{data}$, i.e., α = 0). This allows the model to learn a strong, stable, and accurate mapping from the input parameters to the displacement field without any interference from the physics loss.

2. **Stage 2: Physics-Informed Fine-tuning.** The weights of the converged, pre-trained model are then loaded. In this second stage, the physics loss term is introduced. The weight α is not fixed but is **annealed**—it is gradually increased from 0 to its final target value over a set number of epochs.

This fine-tuning approach proved to be critical for success. By starting from a model that already provides a very good solution, the physics loss acts as a gentle regularizer, "nudging" the pre-trained solution into a nearby region of the parameter space that better conforms to the Navier-Cauchy equations. This prevents the gradient conflicts observed in the naive approach and leads to a stable decrease in both training and validation loss, ultimately yielding a more accurate and physically plausible final model.

## 3.6. Evaluation and Benchmarking

To provide a comprehensive and rigorous assessment of our models, we evaluated their performance from two critical perspectives: predictive accuracy and computational efficiency. All evaluations were performed on a held-out, unseen test set, ensuring an unbiased measure of each model's generalization capabilities.

### 3.6.1. Predictive Accuracy Metrics

We used a suite of three metrics to quantify the accuracy of the predicted displacement fields against the ground truth FEA results.

- **Mean Absolute Error (MAE):** This metric provides a direct, interpretable measure of the average pointwise error in physical units. It is calculated as the mean of the absolute differences between the predicted displacement vectors ($u_{pred}$) and the ground truth vectors ($u_{true}$) over all *N* nodes in a sample:

$$\text{MAE} = \frac{1}{N}\sum_{i=1}^{N}|\hat{u}_i - u_i|$$

  The final MAE reported is the average over all samples in the test set, with units of millimetres (mm).

- **Relative L2 Error (%):** To provide a normalized measure of error that is independent of the absolute magnitude of displacement, we use the Relative L2 Error. This is particularly important for comparing performance across different load cases and signal strengths (e.g., low-force vs. high-force scenarios). It is calculated as the ratio of the L2 norm of the error field to the L2 norm of the ground truth field, expressed as a percentage:

$$\text{Relative } L_2 \text{ Error } = \frac{||u_{\text{pred}} - u_{\text{true}}||_2}{||u_{\text{true}}||_2} \times 100$$

  A lower percentage indicates a more accurate field-level prediction.

- **R² Score (Coefficient of Determination):** A standard statistical measure, the R² score represents the proportion of the variance in the ground truth data that is predictable from the model's predictions. It provides a valuable assessment of the model's goodness of fit. An R² score of 1.0 indicates a perfect fit, while a score of 0 indicates the model performs no better than a constant baseline predicting the mean of the data.

While we focus our evaluation on these full-field metrics, their high fidelity directly implies accuracy on derived Quantities of Interest (QoI), such as maximum deflection, as these are direct functions of the predicted field.

### 3.6.2. Computational Performance Metrics

For surrogate models to be practical, they must offer a significant speed advantage over the original solver. We use two key metrics to quantify this efficiency.

- **Inference Time (ms):** This is the wall-clock time required for a trained model to perform a single forward pass and generate a prediction for one sample from the test set. The reported time is averaged over the entire test set to ensure a stable measurement. This metric directly quantifies the speedup of the surrogate compared to the minutes or hours required for a single FEA simulation.

- **Model Complexity (Parameters, M):** The number of trainable parameters in a model serves as a direct proxy for its size and memory footprint. Reported in millions (M), this metric is crucial for understanding the trade-off between model accuracy and its computational requirements for both training and deployment, especially in resource-constrained environments.

### 3.6.3. Experimental Setup

All models were trained and evaluated using a consistent experimental setup to ensure fair and reproducible comparisons.

- **Data Split:** The datasets were split into training (80%), validation (10%), and test (10%) sets. The validation set was used for hyperparameter tuning and to monitor for overfitting during training, while the test set was strictly held out and used only for the final performance evaluation reported in our results.

- **Hardware:** All training and inference benchmarks were conducted on a consistent hardware platform, specifically using an NVIDIA GeForce RTX 4050 with 6 GB of VRAM.

## 4. Experiments and Results

An exhaustive set of experiments were performed to a evaluate different architectures against the metrics listed in section 3.6. All the results are summarised together in Table 2. This table acts as the primary reference for further analyses in next subsections.

### 4.1. Architectural Showdown: Mesh-Based GNNs vs. Grid-Based U-Nets

Mesh based GNN models are consistently outperforming the grid-based U Net models. On the challenging Low Signal (Generalist) task, the U-Net models, regardless of resolution or the inclusion of attention mechanisms, performed poorly, yielding a Relative L2 Error of over 25%. While the worst GNN model (GCN) performance is far better than the UNet, scoring of 9.7% relative L2 error. The best performers among GNN Graph transformers yields only 3.8% relative L2 error. This means the best among GNNs is nearly seven times more accurate than its best U Net counterpart.

This result can be attributed to "free added learning" in GNNs. They inherit the mesh structure from the FEA model directly. Thus, GNNs should be a natural choice for deep learning-based surrogate models.

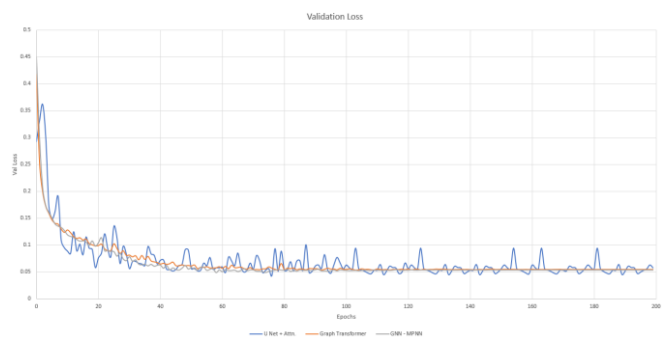

*Figure 4: Validation Loss Curves. The validation loss (Relative L2 Error) over training epochs for the top-performing GNNs (MPNN, Graph Transformer) and the U-Net. The U Net shows a fluctuating validation loss, while the GNNs converges smoothly to lower error, demonstrating their superior learning capability for this problem.*

**Table 2: Comprehensive Evaluation Results**

| Model | Task | MAE (mm) ↓ | R-L2 (%) ↓ | R² Score ↑ | Inference (ms) ↓ | Params (M) ↓ |
|---|---|---|---|---|---|---|
| *LOW SIGNAL (Generalist)* | | | | | | |
| GCN | Generalist | 0.0002 | 9.7473 | 0.9899 | 1.2365 | 0.0521 |
| GAT | Generalist | 0.0002 | 6.8524 | 0.9950 | 0.9444 | 0.3992 |
| MPNN | Generalist | 0.0001 | 3.8348 | 0.9984 | 0.1441 | 0.2990 |
| **MPNN-PINN** | **Generalist** | **0.0001** | **3.6751** | **0.9986** | **0.1667** | **0.2990** |
| Graph Transformer | Generalist | 0.0001 | 3.8524 | 0.9984 | 0.2903 | 1.5800 |
| U-Net (Low-Res) | Generalist | 0.0000 | 25.4867 | 0.9244 | 1.4520 | 5.6107 |
| U-Net (High-Res) | Generalist | 0.0000 | 26.2694 | 0.9194 | 1.8244 | 5.6107 |
| U-Net + Attn | Generalist | 0.0000 | 25.8114 | 0.9219 | 2.4954 | 5.6243 |
| *LOW SIGNAL (Specialist)* | | | | | | |
| GCN | Specialist | 0.0001 | 9.1548 | 0.9880 | 0.7837 | 0.0518 |
| MPNN | Specialist | 0.0001 | 4.0021 | 0.9977 | 0.1726 | 0.2988 |
| GAT | Specialist | 0.0001 | 5.8954 | 0.9950 | 2.3798 | 0.3990 |
| *HIGH SIGNAL (Specialist)* | | | | | | |
| GCN | Specialist | 0.0128 | 8.7901 | 0.9889 | 0.6130 | 0.0518 |
| GAT | Specialist | 0.0089 | 5.4931 | 0.9957 | 2.6013 | 0.3990 |
| MPNN | Specialist | 0.0063 | 4.0354 | 0.9977 | 0.1627 | 0.2988 |
| **MPNN-PINN** | **Specialist** | **0.0055** | **3.5789** | **0.9982** | **0.2751** | **0.2988** |
| **Graph Transformer** | **Specialist** | **0.0042** | **2.6466** | **0.9990** | **0.2237** | **1.5798** |
| U-Net + Attn (High-Res) | Specialist | 0.0001 | 13.0801 | 0.9656 | 1.7742 | 5.6243 |

### 4.2. Performance Hierarchy of GNN Architectures

Among the GNNs, Table 2 clearly shows that models with better expressive complex architecture such as MPNNs and graph transformer perform better than the less complex models like GCN and GAT. The trend can be seen in the generalist models. GCN is the worst model with isotopic averaging resulting in 9.7% R-L2 error. GAT model performs better with 6.8% R-L2 error. The better performance comes from anisotropic weighing of different connections. There is a significant leap in performance in MPNN with only 3.8% R-L2 error. The MPNN model represents the local physics better with its advanced message passing mechanism. Graph transformer model is comparable to the MPNN model. This model learns the global dependencies better.

In the High Signal (Specialist) task, the Graph Transformer achieved an R-L2 error of just 2.65%, while the GCN lagged with an error of 8.79%.

### 4.3. Analysis of Generalist vs. Specialist Models

As discussed earlier in methodologies, we compare "generalist" models to "specialist" models. Multiple loading mechanisms are fed to the generalist model during training. While only single loading mechanism was fed to "specialist" model. Table 3 isolates the best performing generalist and specialist GNN models - MPNN.

**Table 3: Comparison of Generalist vs. Specialist Performance (MPNN, Low Signal)**

| Model | R-L2 (%) ↓ | R² Score ↑ |
|---|---|---|
| MPNN (Generalist) | 3.8348 | 0.9984 |
| MPNN (Specialist) | 4.0021 | 0.9977 |

The MPNN generalist model have 3.8% R-L2 error marginally better than the MPNN specialist model (4%). The results suggest that "generalist" models learn the stress-strain interaction better than the "specialist" models. We hypothesise that multiple load cases act as a kind of multi task learning problem. This could provide better regularization. The value of generalist model is not just better performance. A single generalist model can predict response to any of the different types of loading behaviour. **Figure 5** visually demonstrates this versatility.

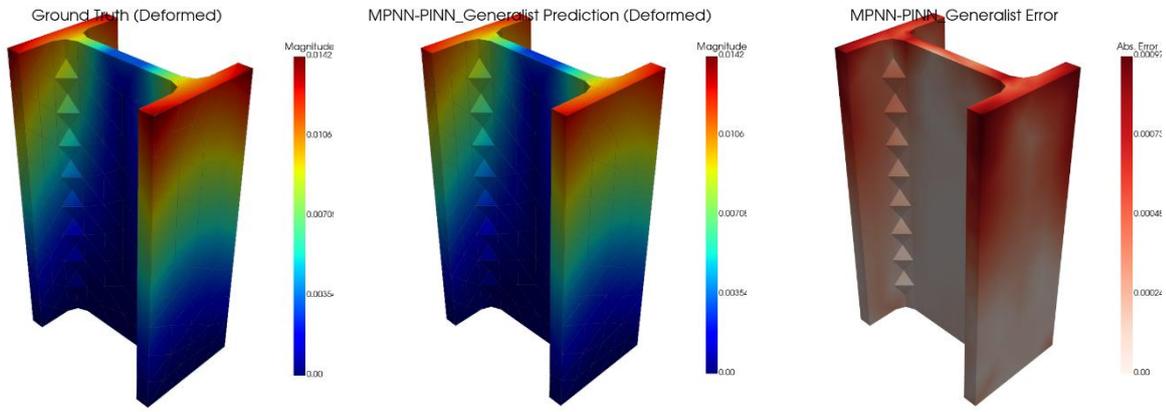

*Figure 5: The Generalist's Versatility.* Predictions from a single **MPNN-PINN (Generalist)** model on three different load cases from the test set. The model correctly captures the distinct deformation physics for vertical bending, horizontal bending, and torsion, confirming its robustness and flexibility.

### 4.3. Efficacy of Physics-Informed Regularization

Table 4 extracts the result for comparison between MPNN model trained on data loss to MPNN model trained on a combination of data and physics loss. It can be clearly seen the physics informed neural networks performed better than the data only models.

**Table 4: Performance Impact of PINN Fine-Tuning on the MPNN Architecture**

| Model | Task | R-L2 (%) ↓ | R² Score ↑ |
|---|---|---|---|
| MPNN | Low Signal (Generalist) | 3.8348 | 0.9984 |
| MPNN-PINN | Low Signal (Generalist) | 3.6751 (-4.2%) | 0.9986 |
| MPNN | High Signal (Specialist) | 4.0354 | 0.9977 |
| MPNN-PINN | High Signal (Specialist) | 3.5789 (-11.3%) | 0.9982 |

The physics integrated MPNN models achieved a lower Relative L2 Error and a higher R² Score for both generalist and specialist datasets. The physics based fine tuning achieved a relative 4.2% lower R-L2 score than data only model on the generalist dataset. This performance improvement is even better on the specialist dataset. The model achieves a 11.3% better R-L2 score than its counterpart. No efforts were made in this study to isolate the root cause for this improvement. This can be attributed to more training time, different activation function (SiLU instead of ReLU) or other trainer parameters. However, we hypothesise that this improvement occurs as we provide physics knowledge directly to the model by adding Navier-Cauchy loss term. This acts as an effective regularizer that guides the model to a physically feasible solution and hiders to goose chase to FEA numerical noise present in ground truth dataset.

The integration of physics loss to the MPNN model was done as part of fine-tuning process with linearly increasing physics loss coefficient to prevent abrupt shock to weight parameters. This approach helped the model to smoothly converge on a plausible solution, see Figure 6 for more details.

### 4.4. The Performance vs. Efficiency Trade-Off

The Graph Transformer model performed the best on high signal data demonstrating best accuracy of 2.65%. However, this accuracy comes with huge computational burden of 1.57M parameters. This model performs task in $O(N^2)$ operations for N nodes. The impact can be seen in inference time; the transformer performs inference in 0.22ms while the MPNN model provides output in only 0.16ms. This improvement may seem trivial in absolute scale, however, we need to consider that the study geometry and physics simulation is relatively simple than most real-world engineering application. Thus, rather than seeing the improvement as 0.06ms, we should note that transformer model is 37.5% slower than the MPNN model.

**Table 5: Performance vs. Efficiency of Top Models (High Signal Specialist Task)**

| Model | R-L2 (%) ↓ | Params (M) ↓ | Inference (ms) ↓ |
|---|---|---|---|
| Graph Transformer | 2.6466 | 1.5798 | 0.2237 |
| MPNN | 4.0354 | 0.2988 | 0.1627 |
| MPNN-PINN | 3.5789 | 0.2988 | 0.2751 |

The MPNN-PINN model, though less accurate than the transformer model, should be considered the best compromise between efficiency and accuracy for applications that require real time monitoring in memory constrained system.

Labels: MPNN - PINN Curriculum Learning
Fine tuning MPNN data only model with PINN Curriculum Learning
MPNN - PINN

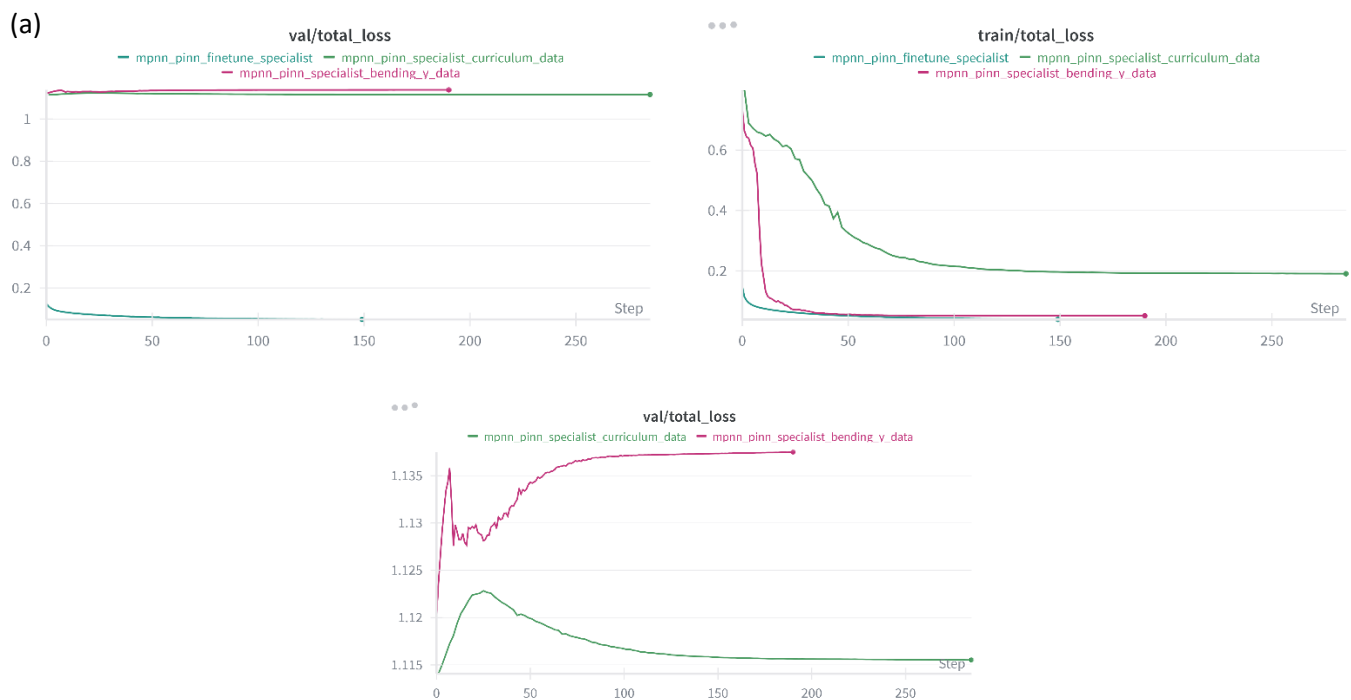

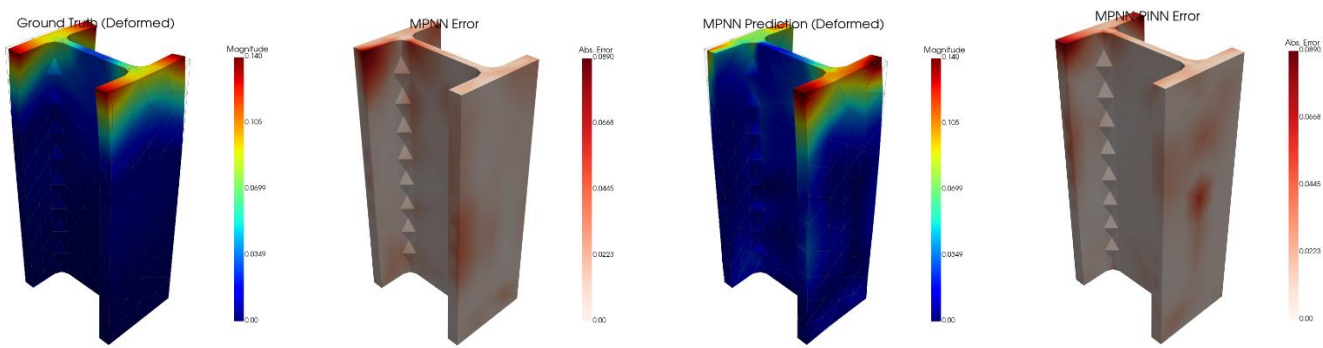

*Figure 6: Training Stability and Qualitative PINN Improvement.* (a) Training curves showing the unstable validation loss of a naive PINN vs. the stable convergence of our curriculum learning strategy. (b) A qualitative comparison of the error maps for the MPNN vs. the MPNN-PINN model on a sample from the test set.

**5. Discussion**

Since the GNN models were constantly outperforming the U Nets, it can be concluded that "inherited" free learning (structure of graph) sets the graph-based model apart from others. As the GNNs operate on native FEA mesh, information loss and approximation errors due to preprocessing is avoided.

Among GNNs more complex message passing methods like MPNN or Graph transformers outshined other architectures. This suggests that stress-strain interaction needs complex mechanism to learn.

Adding the physics knowledge (Navier-Cauchy PDE) to the models, improved the performance of pre-trained data only models. This improvement can be attributed to regularization effect of physics loss. This term hinders the model to chase "numerical" noise in the FEA ground truth data and guides the model to real world physical solutions.

However, adding the physics loss in training makes the model unstable. Deliberate efforts were made to bound the physics loss to 1% to 10% of total loss. Even with constantly lowering the learning rates, adding a learning rate scheduler and a L2 regularization weight decay, the training was unstable. Linear increasing the physics loss coefficient over several epochs (1/4 of total training, 50 epochs for 200 epochs run) helped with stability. But physics integration is best suited as fine tunning a pre-trained data model.

## 6. Conclusion

This work studies development of deep learning surrogate models for finite element analysis. The study presents the method to generate data for I beam structures. Then the study covers development of a range of graph based neural networks like GAT, MCN, MPNNs, transformers and U Net architectures with and without attention mechanism. The study further investigates the impact of inclusion of Navier-Cauchy equations to the performance. We also analysed the trade-off between specialist and generalist models for generalisation.

The study reveals three key findings. First, graph based neural networks outperforms the grid-based U nets for the particular task. This performance gap is more evident in GNNs with complex message passing mechanisms like MPNNs and Graph Transformers. Secondly, integration of fundamental physical laws like Navier-Cauchy PDEs for stress-strain interaction boosts the generalisation. We recommend fine tuning a data only trained model for the integration of PDEs. This greatly helps in training stability. Finally, models when on different kinds of loading mechanism learn the fundamental stress – strain interaction better. These models depict better generalisation.

Even though graph transformers demonstrate best absolute accuracy, MPNN-PINNs (MPNN model fined tuned with physical law integration) closely follows behind. For practical purposes like real time analysis, such as in digital twins with short response time, we should opt for MPPN-PINN over graph transformers because the model has five times lower model parameters than transformer. Thus, they provide a good balance between accuracy and inference speed.

## 7. Limitations and Future Work

Linear elasticity was chosen to be the physical phenomena for this analysis. The work further narrows the scope to I beams. Future study can extend this work to more complicate physical phenomena and diverse geometries including:

- **Non-Linear Physics:** Following the same approach, we can extend the modelling to non-linear materials such as rubbers or silicone and physical non-linearity e.g. plastic deformations.

- **Geometric Generalization:** Further work can work to build a universal surrogate models for beam structures like T bar, L angle C channel etc. This would aim towards better generalization.

- **Advanced PINN Techniques:** Methods such as adaptive sampling, inclusion of explicit boundary condition losses, more advanced curriculum strategies could be explored.

- **Deployment on Engineering Workflows:** Real world value of these surrogate models can be demonstrated when these models can be deployed on downstream tasks such as geometry optimization for specific design problems.

# Appendix

*Hyperparameter Table*

| Category | Parameter | Value | Notes / Component |
|---|---|---|---|
| Training | Optimizer | Adam | torch.optim.Adam |
| | Learning Rate (initial) | 0.0001 | args.lr |
| | Weight Decay | 0.00001 | |
| | Batch Size | 16 | args.batch_size |
| | Epochs | 100 | args.epochs |
| | LR Scheduler | ReduceLROnPlateau | torch.optim.lr_scheduler.ReduceLROnPlateau |
| | Scheduler Patience | 10 epochs | patience=10 |
| | Scheduler Factor | 0.5 | factor=0.5 |
| PINN Framework | PINN Activation | True / False | args.use_pinn |
| | Physics Loss Weight ($\alpha$) | 0.000001 | args.pinn_weight |
| | GNN Activation Function | nn.SiLU() (if PINN) / nn.ReLU() (if not) | Handled in GNN_Base class |
| GNN Architecture | Hidden Feature Dimension | 128 | args.hidden_size |
| | Input Features (Generalist) | 16 | pos(3) + params(10) + one-hot(3) |
| | Input Features (Specialist) | 14 | pos(3) + params(11) |
| | GCN Layers | 3 | GCN_Surrogate |
| | GAT Layers | 3 | GAT_Surrogate |
| | GAT Heads | 4 (for first 2 layers), 1 (for last layer) | GAT_Surrogate |
| | MPNN Layers (MetaLayer) | 3 | MPNN_Surrogate |
| | Graph Transformer Layers | 3 | GraphTransformer_Surrogate |
| | Graph Transformer Heads | 4 (for first 2 layers), 1 (for last layer) | GraphTransformer_Surrogate |
| U-Net Architecture | UNet3D (Small) | | |
| | Initial Channels | 32 | UNet3D_Small class |
| | Channel Progression | 32 → 64 → 128 → 256 | Encoder path |
| | UNet3D (Standard) | | |
| | Initial Channels | 64 | UNet3D class |
| | Channel Progression | 64 → 128 → 256 → 512 | Encoder path |
| | Convolution Kernel Size | 3x3x3 | DoubleConv3D |
| | Upsampling Method | ConvTranspose3d (Kernel 2, Stride 2) | Up module |
| | Attention Block | Squeeze-and-Excitation (SE_Block3D) | args.use_attention |
| | SE Reduction Ratio | 16 | SE_Block3D |